\documentclass[11pt,a4paper]{article}

\usepackage[utf8]{inputenc}
\usepackage[T1]{fontenc}
\usepackage{mathptmx}
\usepackage[scaled=0.92]{helvet}
\usepackage[margin=2.5cm]{geometry}
\usepackage{amsmath,amssymb}
\usepackage[numbers,sort&compress]{natbib}
\usepackage{authblk}
\usepackage{booktabs,tabularx}
\usepackage{enumitem}
\usepackage{xcolor}
\usepackage[most]{tcolorbox}
\usepackage{tikz}
\usetikzlibrary{positioning,arrows.meta,fit,calc}
\usepackage[colorlinks=true,linkcolor=blue!50!black,citecolor=blue!50!black,urlcolor=blue!50!black]{hyperref}

\usetikzlibrary{positioning,arrows.meta,fit,calc,shapes.geometric}

\definecolor{forthblue}{HTML}{0070C0}
\definecolor{forthnavy}{HTML}{002060}
\definecolor{forthgreen}{HTML}{007A37}

\newtcolorbox{statementbox}[1]{enhanced,breakable,colback=forthblue!4,colframe=forthblue!70!black,
  fonttitle=\bfseries,title=#1,left=6pt,right=6pt,top=4pt,bottom=4pt}

\newcommand{\new}[1]{{\color{blue}#1}}

\title{Towards Trustworthy AI Agents through Ontologies, Knowledge Graphs and Agent Memory Management\\
	{\small A Joint Statement from the Olympia AI Data Infrastructure Summit 2026}
}

\author[1,2]{Yannis Tzitzikas\thanks{Coordinating authors. Contact: \texttt{tzitzik@ics.forth.gr}}}
\author[3]{Nikos Ntarmos\protect\footnotemark[1]}
\author[4]{Nick Bassiliades}
\author[5]{Diego Calvanese}
\author[6]{Georgios Chalkiadakis}
\author[7]{Diego Collarana}
\author[8]{Claudia d'Amato}
\author[1]{Martin Doerr}
\author[9]{Yannis Ioannidis}
\author[10]{Shang-Ling Jui}
\author[11]{Arijit Khan}
\author[2,1]{Kostas Magoutis}
\author[12]{Georgios Paliouras}
\author[3]{Jeff Z. Pan}
\author[4]{Ioannis Pitas}
\author[2,1]{Dimitris Plexousakis}
\author[13]{Joseph Sifakis}
\author[2,1]{Grigoris Tsagkatakis}
\author[4]{Grigoris Tsoumakas}
\author[14]{Panos Vassiliadis}
\author[15]{Bo Zhao}

\affil[1]{Institute of Computer Science, FORTH, Heraklion, Greece}
\affil[2]{University of Crete, Greece}
\affil[3]{Huawei Technologies R\&D (UK) Ltd}
\affil[4]{Aristotle University of Thessaloniki, Greece}
\affil[5]{Free University of Bozen-Bolzano, Italy}
\affil[6]{Technical University of Crete, Greece}
\affil[7]{Fraunhofer FIT, Germany}
\affil[8]{University of Bari Aldo Moro, Italy}
\affil[9]{National and Kapodistrian University of Athens, Greece}
\affil[10]{Consultant, Huawei France, France} 
\affil[11]{Aalborg University, Denmark}
\affil[12]{NCSR ``Demokritos'', Greece}
\affil[13]{Verimag, Universit\'e Grenoble Alpes, France}
\affil[14]{University of Ioannina, Greece}
\affil[15]{Aalto University, Finland}

\date{November 2026}

\begin{document}
\maketitle

\begin{abstract}
AI agents are evolving from conversational assistants into autonomous systems that reason, plan and act on behalf of people and organisations. Their weak points are increasingly found not in the language model itself but around it: in what the agent remembers, in how its knowledge is grounded, in how it shares knowledge with other agents, and in how much we can trust its actions. This white paper is the joint statement of the speakers and panelists of the Olympia AI Data Infrastructure Summit 2026 (Heraklion, Crete, 4--5 November 2026), which brought together researchers in knowledge representation, AI agents and data management, and industrial research and development leaders. We state eleven shared principles, summarise the state of the art on enterprise ontologies, neuro-symbolic knowledge engineering, agent memory, trustworthy autonomy and data infrastructure, introduce a reference view and a minimal formal vocabulary that connects these topics, and formulate 
the open questions and shared intentions that we consider most important.
Our central position is that agent memory is a decision problem over information, not merely a storage problem, and that ontologies and knowledge graphs are best seen as a reliability and structure layer inside hybrid memory architectures.
\end{abstract}


\section{Introduction}

Large language models (LLMs) have made it possible to build software agents that interpret goals expressed in natural language, decompose them into steps, call tools and act on enterprise systems~\cite{yao2023react,sumers2024coala}. As such agents move from demonstrators to products, their limitations become clear: 
from time to time they forget or misremember, they hallucinate~\cite{ji2023hallucination},
they use long contexts unevenly~\cite{liu2024lost}, and they are often inconsistent:
the same question, asked in a different way, may receive different and even
contradictory answers~\cite{elazar2021consistency}; moreover, they offer few
guarantees about the actions they take~\cite{sifakis2018autonomous,harel2020autonomics}.
At the same time, organisations hold decades of investment in structured data, ontologies and knowledge graphs~\cite{hogan2021,pan2017linked}, which encode exactly the kind of explicit, shared and auditable knowledge that agents lack.

The Olympia AI Data Infrastructure Summit 2026, hosted by FORTH in Heraklion, Crete, and supported by Huawei, was organised around this tension and around its motto, \emph{``Data shapes memory, Knowledge drives intelligence.''} Its three themes were (i)~agent memory infrastructure, (ii)~enterprise ontology, and (iii)~the future of AI data infrastructure. Over two days, three keynotes~\cite{t_sifakis,t_calvanese,t_ioannidis}, fifteen invited talks~\cite{t_pan,t_doerr,t_zhao,t_collarana,t_bassiliades,t_damato,t_khan,t_tsagkatakis,t_paliouras,t_vassiliadis,t_pitas,t_chalkiadakis,t_tsoumakas,t_magoutis,t_jui} and a panel~\cite{t_panel} addressed one central question:

\begin{quote}
\emph{How can ontologies and structured knowledge become a practical foundation for AI agents that remember, reason, and act reliably?}
\end{quote}

The question deliberately connects three communities, namely ontology and knowledge-representation researchers, AI-agent researchers, and industrial R\&D leaders, and asks to move beyond the general claim that knowledge graphs improve AI: what do ontologies actually contribute to agent memory, what can be implemented today, and what is still missing for dependable products?

At the closing session, participants agreed to write a short white paper that records the shared view of the summit. This document is that paper. Section~\ref{sec:statement} states our principles and shared intentions. Section~\ref{sec:view} introduces a reference view and a minimal vocabulary. Sections~\ref{sec:onto}--\ref{sec:infra} review the state of the art and the work presented at the summit. 
Section~\ref{sec:panel} records the questions debated in the panel, 
Section~\ref{sec:questions} lists open questions, and finally Section~\ref{sec:CR} concludes.

\section{The Olympia 2026 Statement}
\label{sec:statement}

\begin{statementbox}{Principles}
We, the speakers and panelists of the Olympia AI Data Infrastructure Summit 2026, share the following view.
\begin{enumerate}[leftmargin=*,itemsep=2pt,label=\textbf{P\arabic*.}]
\item \textbf{Agent memory is a decision problem, not only a storage problem.} An agent must decide what to preserve, how to represent it, when to update it, what evidence to retrieve, and when not to answer. These choices are coupled: a poor extraction can corrupt a graph, an over-compressed summary can erase a temporal exception, and a highly relevant retrieval can still be insufficient evidence for an action.
\item \textbf{Trust is engineered, not assumed.} Autonomy is defined by functionality, not by a specific technique; trustworthy agents require rigorous design, verification and evaluation around the model.

\item \textbf{State and memory are infrastructure.} Context, experience and reasoning traces need managed, governed and costed storage, like any other enterprise data.
\item \textbf{Neural and symbolic methods are complementary, in both directions.} 
LLMs help build and align semantic knowledge and complete knowledge graphs;
ontologies and reasoning guide, validate and explain the neural components.
\item \textbf{Semantics bridge data and decisions.} Ontologies and knowledge graphs provide stable identity, typed relations, temporal validity, provenance, constraint checking and auditable updates. They are best treated as a reliability and structure layer inside hybrid memory architectures, not as a complete memory solution.
\item \textbf{Agents are the new users of data infrastructure.} Data systems must serve many agents that query, update and remember in parallel, at enterprise scale.
\item \textbf{Learning becomes collaborative and continual.} Agents learn from experience and from each other, and must do so without forgetting what they already know.
\item \textbf{Uncertainty and assurance must be explicit.} Each capability should state which level of assurance it offers (statistical, semantic or formal) and where its limits lie.
\item \textbf{Evaluation is a research problem in itself.} Shared benchmarks, explanation protocols and methods for locating misalignment are needed for memory, ontologies and agents.
\item \textbf{Knowledge lifecycles become continuous.} Provenance-aware, agent-assisted curation keeps knowledge current as the world, and the organisation, change.

\item \textbf{Humans stay in the loop and in control.} 
Agents must keep humans in the loop: people should be able to supervise,
correct and override agents, especially for consequential actions.

\end{enumerate}
\end{statementbox}

\begin{statementbox}{Shared Intentions}
Therefore, we consider it worthwhile to:
\begin{enumerate}[leftmargin=*,itemsep=2pt,label=\textbf{C\arabic*.}]
\item work towards a common reference architecture and vocabulary for agent memory and enterprise ontologies, starting from Section~\ref{sec:view} of this paper;
\item build shared benchmarks and evaluation protocols for agent memory and ontology quality, open to the community;
\item create open testbeds with realistic enterprise scenarios and datasets, together with industrial partners;
\item pursue joint research and innovation projects between academia and industry;
\item organise small working groups per topic, and meet again to review progress.
\end{enumerate}
\end{statementbox}

\section{A Reference View and a Common Vocabulary}
\label{sec:view}

\subsection{Reference view of an enterprise agent}

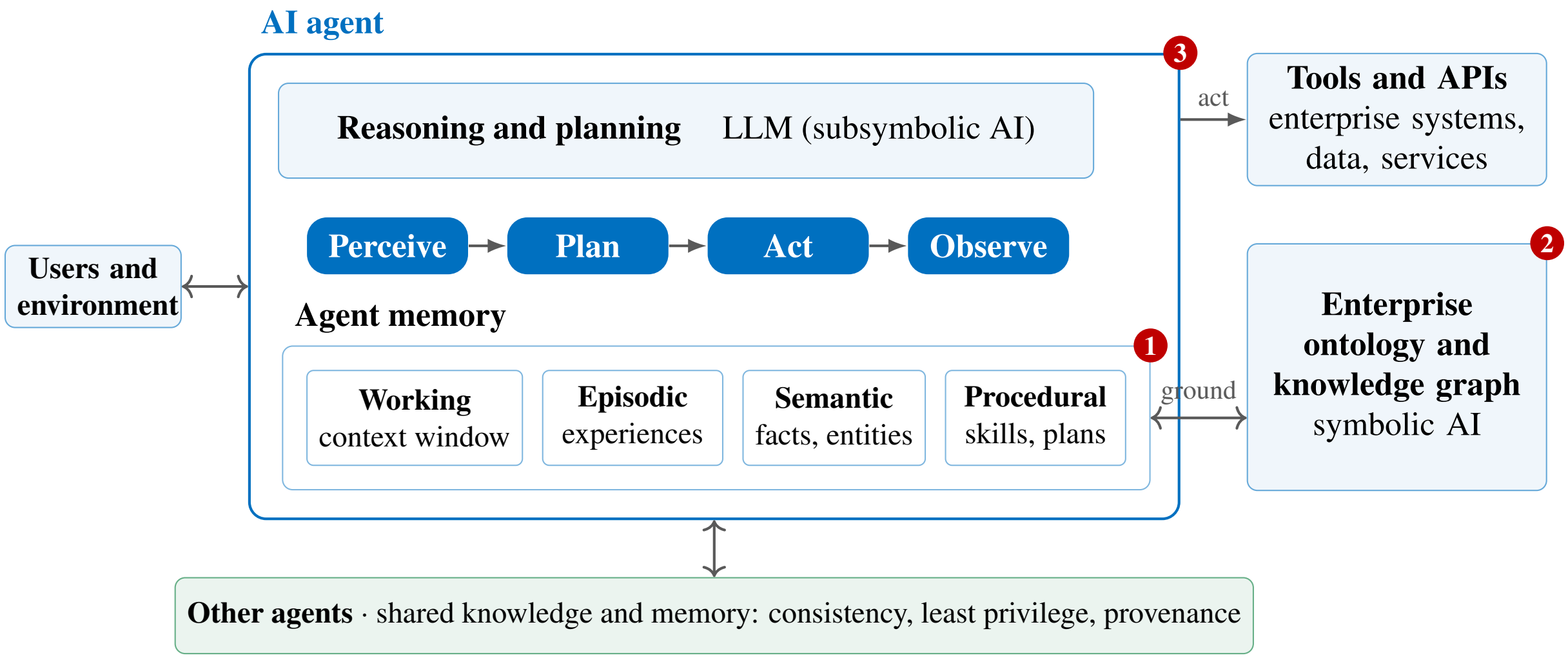
\begin{figure}[t]
\centering
\resizebox{\linewidth}{!}{%
\begin{tikzpicture}[
  font=\small,
  box/.style={draw=forthblue!60, rounded corners=3pt, fill=forthblue!6, align=center, minimum height=8mm},
  chip/.style={rounded corners=6pt, fill=forthblue, text=white, font=\small\bfseries, minimum width=17mm, minimum height=6mm},
  mem/.style={draw=forthblue!50, rounded corners=2pt, fill=white, align=center, minimum width=19mm, minimum height=10mm, font=\footnotesize},
  arr/.style={-{Latex[length=2mm]}, thick, gray!70!black},
  num/.style={circle, fill=red!75!black, text=white, font=\scriptsize\bfseries, inner sep=1.2pt}
]
\node[box, minimum width=86mm, minimum height=10mm] (llm) {\textbf{Reasoning and planning} \quad LLM (subsymbolic AI)};
\node[chip, below=4mm of llm.south west, anchor=north west, xshift=3mm] (p1) {Perceive};
\node[chip, right=4mm of p1] (p2) {Plan};
\node[chip, right=4mm of p2] (p3) {Act};
\node[chip, right=4mm of p3] (p4) {Observe};
\draw[arr] (p1) -- (p2); \draw[arr] (p2) -- (p3); \draw[arr] (p3) -- (p4);
\node[mem, below=10mm of p1.south west, anchor=north west] (m1) {\textbf{Working}\\context window};
\node[mem, right=2mm of m1] (m2) {\textbf{Episodic}\\experiences};
\node[mem, right=2mm of m2] (m3) {\textbf{Semantic}\\facts, entities};
\node[mem, right=2mm of m3] (m4) {\textbf{Procedural}\\skills, plans};
\node[draw=forthblue!50, rounded corners=3pt, fit=(m1)(m4), inner sep=2.5mm, label={[font=\small\bfseries,anchor=south west]north west:Agent memory}] (mem) {};
\node[draw=forthblue, thick, rounded corners=5pt, fit=(llm)(p1)(p4)(mem), inner sep=3mm, label={[font=\small\bfseries,text=forthblue,anchor=south west]north west:AI agent}] (agent) {};
\node[box, right=7mm of agent.north east, anchor=north west, text width=29mm] (tools) {\textbf{Tools and APIs}\\enterprise systems,\\data, services};
\node[box, below=6mm of tools, text width=29mm, minimum height=26mm] (kg) {\textbf{Enterprise}\\\textbf{ontology and}\\\textbf{knowledge graph}\\symbolic AI};
\node[box, left=7mm of agent.west, text width=16mm, font=\footnotesize] (user) {\textbf{Users and}\\\textbf{environment}};
\node[box, below=6mm of agent.south, minimum width=86mm, font=\footnotesize, fill=forthgreen!8, draw=forthgreen!60] (shared) {\textbf{Other agents} $\cdot$ shared knowledge and memory: consistency, least privilege, provenance};
\draw[arr, <->] (user) -- (agent);
\draw[arr] (agent.east |- tools) -- node[above,font=\scriptsize]{act} (tools);
\draw[arr, <->] (mem.east) -- node[above,font=\scriptsize]{ground} (kg.west |- mem.east);
\draw[arr, <->] (agent) -- (shared);
\node[num] at (mem.north east) {1};
\node[num] at (kg.north east) {2};
\node[num] at (agent.north east) {3};
\end{tikzpicture}}
\caption{Reference view of an enterprise AI agent. The markers show where the panel questions sit (Section~\ref{sec:panel}): (1)~what the agent remembers, (2)~keeping memory consistent and trustworthy with ontologies, reasoning and provenance, (3)~the architecture as a whole. The fourth question concerns how all of this becomes a product.}
\label{fig:agent}
\end{figure}

Figure~\ref{fig:agent} shows the reference view used throughout the summit. An LLM
performs reasoning and planning, the \emph{subsymbolic} component, inside a
perceive--plan--act--observe loop~\cite{yao2023react}. The agent uses tools and
enterprise systems and maintains several kinds of memory: following the
cognitive-architecture view of language agents~\cite{sumers2024coala}, we
distinguish \emph{working} memory (the context window), \emph{episodic} memory
(experiences and reasoning traces), \emph{semantic} memory (facts and entities) and
\emph{procedural} memory (skills and plans). Semantic memory is grounded in the
\emph{enterprise ontology and knowledge graph}, the \emph{symbolic}
component~\cite{hogan2021,pan2023tgdk}. The knowledge graph, however, is only
part of the picture: while it provides a persistent and shared representation of
facts, agent memory also encompasses episodic experiences, procedural knowledge,
the working context and other task-dependent state needed for reasoning and
action. Finally, the agent shares knowledge and memory with other agents and with
people.

\subsection{A minimal formal vocabulary}
\label{sec:formal}

To keep the discussion concrete, we fix a small vocabulary. Its purpose is common understanding, not a complete semantics.

\paragraph{Ontology and knowledge graph.} An ontology is a triple $O=\langle C,R,A\rangle$ of classes $C$, relations $R$ and axioms $A$. A knowledge graph is a set of facts $G\subseteq I\times R\times I$ over individuals $I$, such that $O\cup G$ is consistent.

\paragraph{What an agent remembers.} The memory of an agent at time $t$ is $M_t=\langle W,E,S,P\rangle$ (working, episodic, semantic, procedural), with semantic memory grounded in the graph, $S\subseteq G$. An item $x$ is retained only if it is worth its cost:
\begin{equation}
\mathit{keep}(x)\iff \mathit{value}(x)-\mathit{cost}(x)>\theta .
\label{eq:keep}
\end{equation}

\paragraph{Consistent and trustworthy memory.} An update $\Delta$ yields $M_{t+1}=M_t\oplus\Delta$ and is accepted only if $O\cup S_{t+1}$ remains consistent. 
However, logical consistency alone does not establish factual correctness or action acceptability; these also require applicable evidence, version checks, and task-specific constraints \cite{jui2026lpvi}.
Every memory item carries provenance $\mathit{prov}(x)=\langle \mathit{source},\mathit{time}\rangle$~\cite{prov2013}, and a low confidence triggers retrieval of evidence instead of reliance on memory,
i.e., the agent doesn’t blindly trust everything it remembers:
\begin{equation}
\mathit{retrieve}(x) \text{ if } \mathit{conf}(x)<\tau .
\label{eq:retrieve}
\end{equation}
Note that this resembles  cache validity: a low-confidence memory is like a possibly stale cache entry, and retrieving evidence is the equivalent of revalidating it at the source.

\paragraph{Architecture.} Given an observation $o_t$, the agent's policy $\pi$ chooses an action using what it recalls, $a_t=\pi\big(o_t,\mathit{recall}(M_t,o_t)\big)$. Questions $q$ over enterprise data $D$ are answered through the ontology and mappings, as \emph{certain answers}~\cite{poggi2008,xiao2018}:
\begin{equation}
\mathit{ans}(q)=\mathit{cert}\big(q,\langle O,\mathit{mappings},D\rangle\big).
\label{eq:obda}
\end{equation}

\paragraph{From research to product.} An organisation adopts an ontology-based approach when $\mathit{gain}(\mathit{quality})>\mathit{cost}(\mathit{build}+\mathit{maintain})$.

\section{Enterprise Ontologies and Knowledge Graphs}
\label{sec:onto}

\paragraph{State of the art.} Ontologies, as explicit specifications of a shared conceptualisation~\cite{gruber1993}, have long been used to integrate heterogeneous information, and knowledge graphs~\cite{hogan2021} made the approach mainstream in industry~\cite{pan2017linked}. Long-lived standards show both the value and the cost of a carefully designed core ontology: the CIDOC Conceptual Reference Model (ISO~21127) is an event-centric ontology for cultural heritage information maintained for more than two decades~\cite{doerr2003}. 
Mappings of existing sources to ontologies can be leveraged for building Materialized Knowledge Graphs \cite{marketakis2017x3ml}, or Virtual Knowledge Graphs \cite{calvanese2017ontop}. 
The benefits of the materialized approach are:
(a) flexibility in transformation logic (including ability to curate and fix problems), (b) faster responses (in query answering but also in other tasks, e.g., if one wants to use it for applying an entity matching   technique), (c) decoupling of access load  from the underlying sources, and (d) decoupling of the release management of the integrated  resource from the management cycles of the underlying sources.
However, the materialized approach requires 
paying the cost for hosting the warehouse, and refreshing it  periodically \cite{marketakis2017x3ml,t_doerr}.
On the other hand, one advantage (but in some cases disadvantage) of virtual integration is the real-time reflection of source updates in integrated access. In general, the higher complexity of the system (and the quality of service demands on the sources) is only justified if immediate access to updates is indeed required.
In general, hybrid approaches are required.

Integrating many independently created datasets requires instance matching, link discovery and services for dataset search and integrated querying~\cite{mountantonakis2019,collarana2017minte}, 
however AI-assisted methods could greatly reduce the required cost (e.g.,  \cite{marketakis2026exploiting}).
Moreover  querying knowledge graphs at scale raises its own challenges~\cite{khan2023kgquerying}. 
Exploratory access methods such as faceted search help users who do not know the schema~\cite{tzitzikas2017faceted}. Knowledge representation research on temporal knowledge, action theories and belief revision provides the formal tools to describe how a knowledge base should change~\cite{agm1985,flouris2008}.

\paragraph{At the summit.} Doerr discussed knowledge about the past, which rests on evidence and interpretation; event-centric models with provenance make it integrable \cite{Doerr2025IdentifiableIndividuals}, and LLMs now make it possible to create very large knowledge bases~\cite{t_doerr}. Collarana argued that knowledge graphs give agents a world model with semantics, stable identifiers, provenance, constraints and governance, and that agents can make costly curation continuous~\cite{t_collarana}. Calvanese's keynote presented knowledge graphs from an industrial perspective, where virtual knowledge graphs let the ontology reach data where it lives~\cite{t_calvanese}.
In enterprise settings, these capabilities can improve auditability, provenance tracking, regulatory compliance and the consistency of decisions taken across agents and human operators.

\paragraph{Challenges.} Enterprises must decide how broad an ontology should be, who is accountable for its correctness, and how to keep mappings, vocabularies and links current as sources change; they rarely have the governance structures of standardisation bodies.

\section{Neuro-Symbolic Knowledge Engineering}
\label{sec:ns}

\paragraph{State of the art.} The relationship between LLMs and knowledge graphs is one of the most active research areas~\cite{pan2023tgdk,pan2024roadmap}. Knowledge graph embeddings~\cite{bordes2013,damato2023machine} scale to large graphs but mostly ignore schema and reasoning, and machine learning for the Semantic Web has highlighted the need to combine both views~\cite{damato2020}. In the other direction, LLMs can draft ontologies, suggest alignments, extract facts and translate questions into structured queries; for event-centric ontologies, prompting with ontology path patterns improves query generation~\cite{mountantonakis2024text2cidoc}. Ontology change, including evolution, versioning and debugging, has been studied as a family of problems~\cite{flouris2008}, and LLM-driven evolution revives it at a new scale.

\paragraph{At the summit.} 
Bassiliades presented neuro-symbolic approaches in which LLM-generated ontology drafts are validated and refined by symbolic reasoning 
\cite{Skaperdas2026OGNSD}, while ontological constraints guide knowledge-graph completion \cite{Skaperdas2026EurostatKG}, as discussed in his summit talk ~\cite{t_bassiliades}.
d’Amato showed that most of the embedding-based link prediction methods ignore the schema,
that common benchmarks discard it, and that explanations need a unified evaluation protocol~\cite{t_damato}.
Pan discussed the shift from scaling model parameters to scaling agent systems,
arguing that larger models do not remove the long-tail knowledge problem and that
longer contexts are not enough, since relevant evidence gets lost in long
prompts~\cite{liu2024lost}; instead, agents need graph-based retrieval,
structured task state, ontology-guided reasoning and coordination, and must scale
along memory, state and collaboration as well~\cite{t_pan,pan2023tgdk}.

\paragraph{Challenges.} We lack agreed measures of the quality and business value of an ontology, methods to explain automated changes to their owners, and benchmarks that keep the schema.

\section{Agent Memory and State}
\label{sec:memory}

\paragraph{State of the art.} Retrieval-augmented generation treats an external corpus as non-parametric memory~\cite{lewis2020rag}, and graph-based variants exploit structure for multi-hop questions and summarisation~\cite{edge2024graphrag}. Agent architectures increasingly manage memory explicitly: MemGPT pages information between a limited context and external storage~\cite{packer2023memgpt}, and generative agents store, retrieve and reflect on experiences~\cite{park2023generative}. Larger context windows do not remove the need, since models use long contexts unevenly~\cite{liu2024lost}. At the serving layer, memory management determines the efficiency of LLM inference~\cite{kwon2023vllm}.

\paragraph{At the summit.} 
Zhao identified a ``state wall'' between evolving agentic state
and runtime decisions~\cite{t_zhao}. His related work addresses complementary
aspects of efficient state management and adaptive execution:
cost-aware reduction of shared intermediate state for sequential
pattern matching, with relevance to applications such as GraphRAG
\cite{yu2026sharp}; transformations of training states to enable
dynamic parallelism in deep learning through parallelizable tensor
collections \cite{wagenlander2024tenplex}; and adaptive parallelism
and experience dispatch for agentic reinforcement learning under
dynamically changing context lengths \cite{tan2026earl}.
Khan argued that the AI harness (retrieval, tools, memory and orchestration) largely decides reliability, and presented knowledge-graph-based retrieval for matching, alignment and fact checking~\cite{t_khan}. Magoutis showed that memory is central to serving LLMs and that scaling AI services means moving state across nodes~\cite{t_magoutis}. 
Shang-Ling Jui presented Learn–Preserve–Verify–Interpret (LPVI), distinguishing candidate generation, preservation of specified constraints and semantics, independent verification, and interpretation for engineering decisions ~\cite{t_jui}. The framework emphasizes that guarantees depend on explicit assumptions and that evidence must remain applicable to the artifact and specification versions concerned.

\paragraph{Challenges.} The value of a memory item, Eq.~(\ref{eq:keep}), is only known in the future, while its costs are paid now; forgetting, decay and revision are equally hard. Belief revision~\cite{agm1985} offers principled starting points that have not yet been connected to LLM agents.
In general, agent memory maintenance mirrors cache admission and replacement, where the optimal eviction decision depends on future accesses; agent memory adds semantic dependencies between items, so evicting or abstracting one item may invalidate others.

\section{Trust, Verification and Safe Evolution}
\label{sec:trust}

\paragraph{State of the art.} Autonomy should be characterised by functionality rather than by techniques, and trustworthy autonomous systems require far more than machine learning~\cite{sifakis2018autonomous,harel2020autonomics}. Model checking provides strong guarantees for systems with explicit models~\cite{clarke2009}, but LLM-based agents lack such models and are prone to hallucination~\cite{ji2023hallucination}. Multi-agent settings add questions of cooperation and incentives~\cite{chalkiadakis2011cgt}.

\paragraph{At the summit.} Sifakis framed autonomy and trust as system-level properties that require rigorous design and validation, and argued that symbolic AI is needed~\cite{t_sifakis}. 
Chalkiadakis presented a novel methodology for uncertainty-driven adaptive exploration, that allows agents to evolve {\em safely} from experience~\cite{bakopoulos2026uncertainty,t_chalkiadakis}.  Shang-Ling Jui distinguished statistical, semantic and formal levels of assurance~\cite{t_jui}.

\paragraph{Challenges.} Assurance must become explicit per capability, and self-evolving agents must preserve the constraints in $O$ and the traces in their episodic memory as they change.

\section{Data Infrastructure, Learning and Applications}
\label{sec:infra}

\paragraph{State of the art.} Decades of data-management research on query optimisation~\cite{ioannidis1996qo}, statistics for estimation~\cite{ioannidis2003histograms}, data integration and ETL~\cite{vassiliadis2002etl,SimitsisSV23},
semantic active caching~\cite{triantafillou2004cache}, scalable query processing~\cite{ntarmos2014rankjoin},
transactions~\cite{gray1992}, access control~\cite{saltzer1975} and constraints~\cite{shacl2017} provide building blocks that agent memory can reuse. Intentional analytics replaces declarative queries with analytical intentions~\cite{vassiliadis2019intentional}. Continual and decentralised learning must avoid catastrophic forgetting~\cite{kirkpatrick2017}. Shared benchmarks have driven progress in semantic indexing and question answering~\cite{tsatsaronis2015}, where multi-label classification plays a central role~\cite{tsoumakas2007multilabel}, and machine learning for scientific data provides demanding domains~\cite{tsagkatakis2019survey}.

\paragraph{At the summit.} Ioannidis's keynote addressed the future of data infrastructure in the agentic era, in which agents become first-class users of data systems and memory, provenance and governance become infrastructure concerns~\cite{t_ioannidis}. Vassiliadis argued that users should state intentions and receive data stories~\cite{t_vassiliadis}. 
Pitas presented decentralised networks of agents that teach and learn from each other without catastrophic forgetting~\cite{t_pitas}. In particular, the Learning-by-Education Node Community (LENC) framework enables agents to exchange and accumulate knowledge through teacher–student interactions, while preserving previously acquired capabilities. This provides an interesting perspective on collaborative agent learning and the mechanisms through which agents can share and retain knowledge across distributed environments.
Tsoumakas discussed efficient relation extraction \cite{tsoumakas2}, 
clinical coding \cite{tsoumakas3}, 
scientific summarization \cite{tsoumakas4}, 
custom retrieval-augmented generation, continual learning and data-science agents 
\cite{tsoumakas1,t_tsoumakas}.
Paliouras presented BioASQ not only as a challenge but as a long-standing
infrastructure and ecosystem for semantic extraction from the biomedical
literature: for more than ten years it has provided curated benchmark datasets,
shared evaluation services and a common methodology, around which a community of
more than 180 teams from 41 countries has formed~\cite{tsatsaronis2015,t_paliouras}.
Its impact extends beyond research: the US National Library of Medicine moved to
fully automated MeSH indexing, showing how a sustained evaluation infrastructure
can carry methods from research into operational use.
Tsagkatakis presented agentic frameworks for space data, from semantic access and agent memory to operational services~\cite{t_tsagkatakis}.

\section{The Panel: Ontology and Agent Memory}
\label{sec:panel}

The panel~\cite{t_panel} brought together complementary perspectives on data infrastructure, reasoning and data access, knowledge representation and evolution, data systems, and learning over knowledge graphs. It addressed four questions.

\begin{enumerate}[leftmargin=*,itemsep=3pt]
\item \textbf{What should an AI agent actually remember?} Today's agents keep summaries, retrieve documents and use knowledge graphs. Which facts, entities, events, preferences, goals, and actions and decisions should be represented explicitly, and where do ontologies make a fundamental difference compared with information-retrieval indexes and vector stores?
\item \textbf{How do we keep agent memory consistent, evolving and trustworthy?} How should agents update memory, handle conflicts and separate facts from assumptions; when should they retrieve evidence rather than trust what they remember (Eq.~\ref{eq:retrieve}); and how can a data system manage memories for hundreds of thousands of agents in parallel?
\item \textbf{What is the right architecture?} Knowledge graph, ontology-based data layer (Eq.~\ref{eq:obda}), vector store, relational database, or a combination; what belongs in the foundation model and what in explicit representations, tools and reasoning services; and how should data be stored and managed for the right trade-offs in latency, scalability, cost and maintainability?
\item \textbf{What would it take to build a product capability within two years?} Which use case should be prioritised, what is the biggest remaining obstacle (knowledge acquisition, ontology construction, integration, reasoning performance, evaluation, privacy, or the cost of memory), and what measurable improvement would convince a product team?
\end{enumerate}


\subsection*{Main points of the discussion}

\paragraph{What should an agent remember?}
Although cognitive science has long distinguished several forms of memory, the
agent community currently focuses mostly on the harness around the model:
retrieval, tools and orchestration. The panel argued that the memory embedded in
a model's parameters is not explainable, whereas symbolic memory is: explainability
therefore requires symbolic representations with clear semantics. Beyond
explainability, symbolic AI offers a persistent semantic layer, a shared point of
reference across agents and people, an internal state that can be queried, and the
ability of an agent to introspect on what it knows. Memory should also support
several levels of abstraction; abstraction can itself be seen as a principled form
of forgetting. Two dimensions of symbolic semantics should not be overlooked: the
temporal dimension (when something held) and the procedural dimension (how things
are done). Since graphs are everywhere, graphs and AI agents meet at many points:
knowledge graphs can help users both before prompting, by structuring the request,
and after it, for example to calibrate answers. Neuro-symbolic combinations can
also counter the overconfidence of neural models, which is a precondition for
trust. Overall, the panel saw agent memory as \emph{multi-modal and multi-layered},
and stressed that the purpose of an agent, and of its memory, should be made
explicit; this calls for goal modelling and for benchmarks that measure whether
memory serves those goals.

\paragraph{Negative knowledge and counterfactual reasoning.}
Another point raised in the panel, which found wide agreement, concerns the role
of \emph{negative knowledge} and \emph{counterfactual reasoning}. Agents need to
represent not only what holds, but also what is known \emph{not} to hold: failed
actions, rejected hypotheses, excluded options, or properties that an entity
explicitly lacks. Most knowledge graphs follow the open-world assumption, in which
missing facts are simply unknown rather than false~\cite{reiter1978cwa}, so such
knowledge must be stated explicitly~\cite{arnaout2021negative}. Logical
formalisations of ontologies make this possible: for instance, the first-order
logic expression of the CIDOC Conceptual Reference Model supports reasoning
with negated properties~\cite{meghini2018fol}. Counterfactual reasoning, that is,
reasoning about what would have happened under different conditions, is equally
central to human cognition and to the way people combine and blend
concepts~\cite{fauconnier2002way}; for agents, it supports planning, explaining
decisions and learning from mistakes.

\paragraph{Consistent, evolving and trustworthy memory.}
Symbolic representations make it possible to separate facts from assumptions, a
distinction that is hard to maintain in purely neural memory. Full consistency,
however, is probably not achievable across all layers of memory, and what counts
as a consistency violation or a conflict may itself depend on the context in which
the knowledge is used.
As agents increasingly share and update common knowledge, mechanisms for provenance-aware conflict resolution, ownership management and coordination of concurrent memory updates become essential infrastructure capabilities.

\paragraph{Architecture.}
On how to store and manage memory for the right trade-offs in latency,
scalability, cost and maintainability, the panel agreed that there is no single
answer: the appropriate combination of knowledge graphs, ontology-based data
access, vector stores and relational databases depends on the use case.

\paragraph{A product within two years.}
The panelists identified medicine, e-science, the environment and decision support
for companies as priority use cases. They proposed pushing computation as close to
the data as possible, both for efficiency and for privacy. Two breakthroughs were
seen as necessary: \emph{operationalising trustworthiness}, so that trust becomes
a measurable and enforceable property rather than an aspiration, and
\emph{blending} the bottom-up approach of large language models with the top-down
approach of symbolic AI.

\section{Open Questions}
\label{sec:questions}

Table~\ref{tab:map} relates the summit contributions to the principles of Section~\ref{sec:statement}. We consider the following questions the most important.

\begin{enumerate}[leftmargin=*,itemsep=2pt,label=\textbf{Q\arabic*.}]
\item How broad should an enterprise ontology be, and how should responsibility for its correctness be shared between domain experts, engineers and agents?
\item What is a meaningful quality measure for an ontology, one that captures business value and not only logical consistency?
\item Which ontology changes can be delegated to LLMs, and how can every accepted change be explained to its owners?
\item How can the value of a memory item be estimated before it is needed, and how should forgetting, decay and revision be implemented?
\item How should semantic memory stay consistent with the enterprise knowledge graph when either changes?
\item Which memory abstractions should data and serving infrastructures expose, so that many agents can work in parallel without losing state?
\item Which levels of assurance can realistically be offered for LLM-based agents, and how should their limits be communicated?
\item How can access policies be defined over semantic memory, and how can conflicting writes by different agents be reconciled?
\item Which shared benchmarks would advance agent memory and ontology quality, in the way BioASQ advanced biomedical semantic indexing?
\item How should explanations, for link predictions as well as for agent actions, be evaluated reproducibly?

\item 


How should humans be kept in the loop in practice: which decisions require
human approval, at what level of detail should agents explain themselves, and
which modes of collaboration and training between AI and domain experts should be
installed for organizing human approval in a scalable way? More broadly, how can
humans act as agents themselves, sharing and enriching knowledge and memory with
AI agents almost as peers, with ontologies and knowledge graphs as persistent
stores of the results of this interaction?

\end{enumerate}

\begin{table}[htbp]
\centering
\small
\caption{Summit contributions and the principles they most directly support. The mapping is indicative.}
\label{tab:map}
\begin{tabularx}{\textwidth}{@{}l>{\raggedright\arraybackslash}Xl@{}}
\toprule
Contributor & Contribution & Principles\\
\midrule
Sifakis & Trustworthy autonomous agents (keynote) & P2, P8, P11 \\
Calvanese & Knowledge graphs from an industrial perspective (keynote) & P5, P6\\
Ioannidis & The future of data infrastructure in the agentic AI era (keynote) & P3, P6\\
Pan & From parameter scaling to agent scaling & P3, P4, P11\\
Doerr & Challenges of knowledge about the past & P5, P10, P11\\
Zhao & State management for agentic AI systems & P1, P3\\
Collarana & Grounding AI agents in enterprise knowledge & P5, P10\\
Bassiliades & Neuro-symbolic ontology engineering and KG completion & P4, P8, P11\\
d'Amato & Neural-symbolic link prediction and explanations & P4, P9, P11\\
Khan & Graph-native LLMs and the AI harness & P2, P3, P9\\
Tsagkatakis & Agentic frameworks for space data management & P3, P6\\
Paliouras & The BioASQ challenge & P9, P11\\
Vassiliadis & From analytic intentions to data stories & P5, P11\\
Pitas & Decentralized machine learning & P7\\
Chalkiadakis & Safe uncertainty-driven adaptive exploration & P7, P8\\
Tsoumakas & From semantic extraction to autonomous agents & P6, P7, P11\\
Magoutis & Autoscaling of AI services & P3, P6\\
Shang-Ling Jui & Ontology, memory and verifiable reasoning & P1, P5, P8, P11\ \\
Plexousakis & Panel: knowledge representation, evolution and reasoning & P1, P10, P11\\
\bottomrule
\end{tabularx}
\end{table}

\begin{figure}[tp]
\centering
\resizebox{\linewidth}{!}{%
\begin{tikzpicture}[
  font=\footnotesize,
  arr/.style={-{Latex[length=2.2mm]}, thick, gray!70!black},
  hdr/.style={font=\small\bfseries, text=white, inner sep=3pt, rounded corners=2pt, minimum height=5mm},
  card/.style={draw, rounded corners=3pt, align=left, inner sep=5pt, text width=#1},
  chip/.style={draw=forthblue!60, fill=forthblue!8, rounded corners=6pt, align=center, inner sep=3pt, font=\footnotesize, minimum height=6mm}
]
\node[card=172mm, draw=forthgreen!70, fill=forthgreen!8, align=center, font=\normalsize] (q) at (0,0)
 {\textbf{Central question:} How can ontologies and structured knowledge become a practical foundation\\for AI agents that remember, reason, and act reliably?};
\node[chip, text width=54mm] (c1) at (-60mm,-13mm) {Knowledge representation and ontologies};
\node[chip, text width=54mm] (c2) at (0,-13mm) {AI-agent research};
\node[chip, text width=54mm] (c3) at (60mm,-13mm) {Data management and industrial R\&D};
\node[card=172mm, draw=forthnavy, fill=forthnavy, text=white, align=center, font=\normalsize] (core) at (0,-28mm)
 {\textbf{Core position:} agent memory is a \emph{decision problem} over information, not only storage;\\ontologies and knowledge graphs are a \emph{reliability and structure layer} inside hybrid memory architectures.};
\draw[arr] (q.south) -- ++(0,-2.5mm);
\foreach \c in {c1,c2,c3} \draw[arr] (\c.south) -- (\c.south |- core.north);
\node[card=52mm, draw=forthblue!70, fill=forthblue!5, anchor=north] (prin) at (-60mm,-45mm) {%
\textbf{\textcolor{forthblue}{Memory}}\\
$\cdot$ a decision problem\\
$\cdot$ state and memory are infrastructure\\
$\cdot$ agents: the new users of data systems\\[2pt]
\textbf{\textcolor{forthblue}{Knowledge}}\\
$\cdot$ neural and symbolic, both directions\\
$\cdot$ semantics bridge data and decisions\\
$\cdot$ continuous knowledge lifecycles\\[2pt]
\textbf{\textcolor{forthblue}{Trust and learning}}\\
$\cdot$ trust is engineered, not assumed\\
$\cdot$ explicit uncertainty and assurance\\
$\cdot$ evaluation is research in itself\\
$\cdot$ collaborative, continual learning\\
$\cdot$ humans stay in the loop};
\node[hdr, fill=forthblue, anchor=south west] at (prin.north west) {Shared principles (Sec.~\ref{sec:statement})};

\node[card=52mm, draw=forthnavy!70, fill=forthnavy!5, anchor=north] (voc) at (0,-45mm) {%
\textbf{\textcolor{forthnavy}{Reference view}}\\
LLM in a perceive--plan--act--observe loop, with tools, memory, an ontology and knowledge graph, and other agents\\[3pt]
\textbf{\textcolor{forthnavy}{Vocabulary}}\\[1pt]
$M_t=\langle W,E,S,P\rangle,\ S\subseteq G$\\[1pt]
$\mathit{keep}(x)\iff \mathit{value}-\mathit{cost}>\theta$\\[1pt]
update accepted iff $O\cup S_{t+1}$ consistent\\[1pt]
$\mathit{retrieve}(x)$ if $\mathit{conf}(x)<\tau$\\[1pt]
$\mathit{ans}(q)=\mathit{cert}(q,\langle O,\mathit{map},D\rangle)$\\[3pt]
\textbf{\textcolor{forthnavy}{Memory as caching}}\\
low confidence $\approx$ stale entry, revalidated at the source; keeping and forgetting $\approx$ admission and eviction};
\node[hdr, fill=forthnavy, anchor=south west] at (voc.north west) {Reference view (Sec.~\ref{sec:view})};

\node[card=52mm, draw=forthgreen!70, fill=forthgreen!5, anchor=north] (pan) at (60mm,-45mm) {%
\textbf{\textcolor{forthgreen}{What to remember}}\\
multi-modal, multi-layer memory; explainability needs symbolic semantics; time and procedures matter; abstraction as forgetting; explicit purpose and goals\\[3pt]
\textbf{\textcolor{forthgreen}{Trustworthy memory}}\\
separate facts from assumptions; consistency depends on context; provenance-aware conflict resolution for concurrent updates; negative knowledge and counterfactuals\\[3pt]
\textbf{\textcolor{forthgreen}{Architecture and product}}\\
no single architecture; compute close to the data; use cases in medicine, e-science, environment, enterprises; operationalise trust; blend bottom-up and top-down};
\node[hdr, fill=forthgreen, anchor=south west] at (pan.north west) {Panel insights (Sec.~\ref{sec:panel})};
\draw[arr] (core.south -| prin.north) -- ++(0,-4mm);
\draw[arr] (core.south) -- ++(0,-4mm);
\draw[arr] (core.south -| pan.north) -- ++(0,-4mm);
\path let \p1=(prin.south), \p2=(voc.south), \p3=(pan.south) in
  node[card=172mm, draw=gray!60, fill=gray!8, align=center, anchor=north] (sota) at (0,{min(\y1,\y2,\y3)-4mm})
 {\textbf{State of the art and summit contributions (Sec.~\ref{sec:onto}--\ref{sec:infra})}\\enterprise ontologies and knowledge graphs $\cdot$ neuro-symbolic knowledge engineering $\cdot$ agent memory and state $\cdot$\\trust, verification and safe evolution $\cdot$ data infrastructure, learning and applications};
\node[card=172mm, draw=red!60!black, fill=red!5, align=center, anchor=north] (oq) at ([yshift=-5mm]sota.south)
 {\textbf{\textcolor{red!60!black}{Open questions (Sec.~\ref{sec:questions}):}} ontology scope and quality $\cdot$ value of memory and forgetting $\cdot$ consistency for many parallel agents $\cdot$\\assurance levels and human oversight $\cdot$ access policies over semantic memory $\cdot$ shared benchmarks and explanation evaluation};
\draw[arr] (sota.south) -- (oq.north);
\foreach \n in {prin,voc,pan} \draw[arr] (\n.south) -- (\n.south |- sota.north);
\node[card=118mm, draw=forthblue!70, fill=forthblue!8, align=center, anchor=north west] (int) at ([yshift=-5mm]oq.south west)
 {\textbf{Shared intentions (Sec.~\ref{sec:statement})}\\reference architecture and vocabulary $\cdot$ shared benchmarks $\cdot$\\open testbeds $\cdot$ joint projects $\cdot$ working groups};
\node[card=46mm, draw=forthgreen, fill=forthgreen, text=white, align=center, anchor=north east, font=\small\bfseries] (goal) at ([yshift=-5mm]oq.south east)
 {Trustworthy AI agents\\that remember, reason\\and act reliably};
\draw[arr] (oq.south -| int.north) -- (int.north);
\draw[arr] (int.east) -- (goal.west);
\end{tikzpicture}}
\caption{The white paper at a glance: from the central question and the core position of the summit, through shared principles, a reference view and vocabulary, panel insights and the state of the art, to open questions and shared intentions towards trustworthy AI agents.}
\label{fig:summary}
\end{figure}
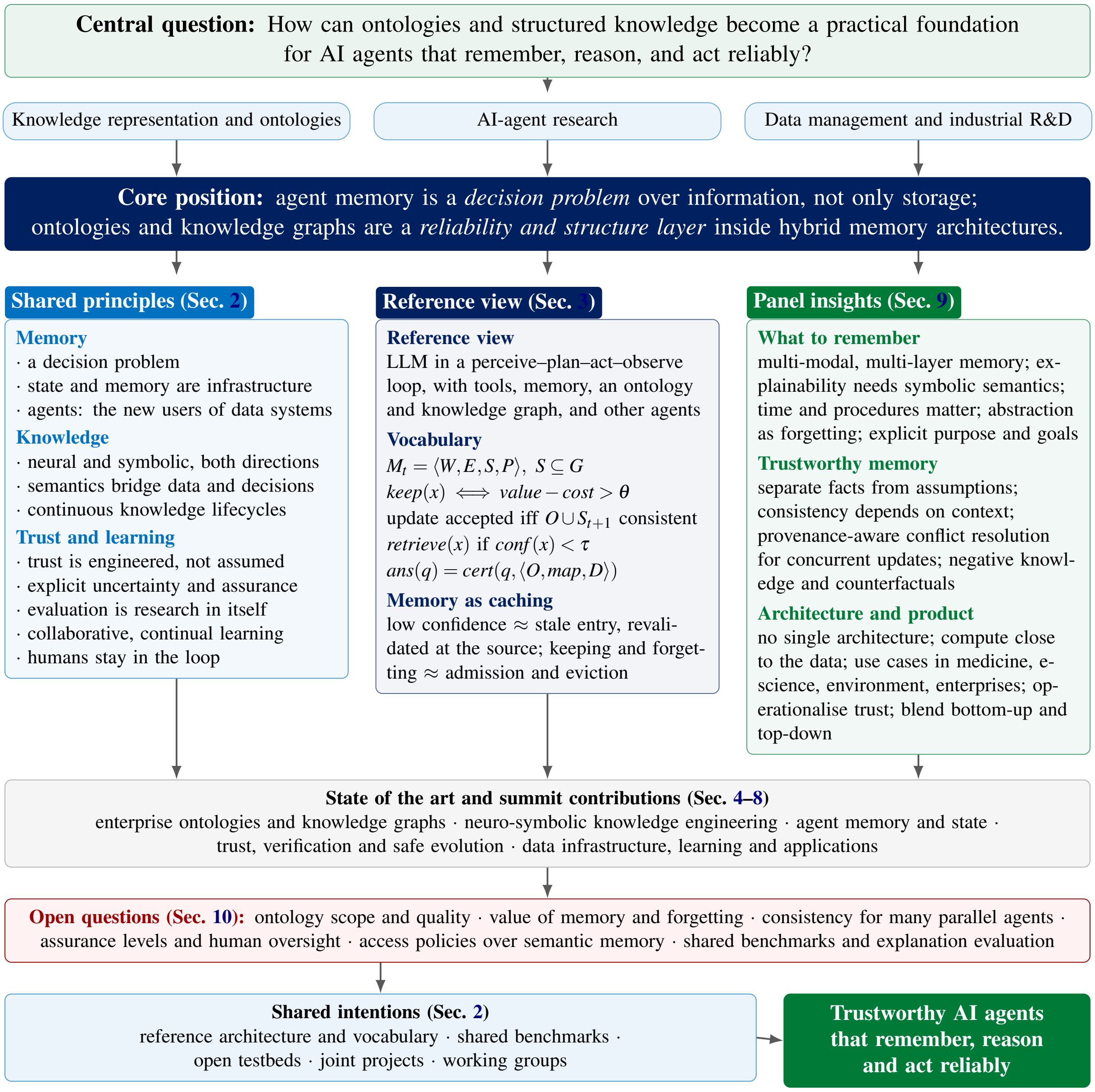

\section{Conclusions}
\label{sec:CR}

The summit showed a remarkable convergence. Researchers in knowledge representation, knowledge graphs, data management, machine learning, distributed systems and formal verification arrived, from different directions, at the same conclusion: dependable AI agents require explicit semantics, explicitly managed memory and explicit assurance. Agent memory is a decision problem over information; ontologies and knowledge graphs are a reliability and structure layer within hybrid architectures; and the central research challenge is to combine learned flexibility with explicit, testable constraints while measuring utility, cost and risk. We invite the wider community to join us in addressing the questions above.
Figure~\ref{fig:summary} summarises the paper at a glance.

\section*{Acknowledgements}
We thank the Foundation for Research and Technology -- Hellas (FORTH) for hosting the summit and Huawei for supporting it, in particular See Yeen Tan, Shen Yan and Yongjian Lyu. We thank Hong Zhou for the opening address, the session chairs Theodore Patkos and Giorgos Flouris, the local organising team at FORTH-ICS, Lida Harami, Maria Moutsaki and Michalis Mountantonakis, and all participants for two days of ideas and dialogue.

\bibliographystyle{plainnat}
\bibliography{olympia2026_statement}

@inproceedings{bakopoulos2026uncertainty,
  author    = {Leonidas Bakopoulos and Georgios Chalkiadakis},
  title     = {A Novel Framework for Uncertainty-Driven Adaptive Exploration},
  booktitle = {Proceedings of the 25th International Conference on Autonomous Agents and Multiagent Systems ({AAMAS} 2026)},
  address   = {Paphos, Cyprus},
  month     = may,
  year      = {2026},
  publisher = {IFAAMAS},
  url       = {https://www.ifaamas.org/Proceedings/aamas2026/pdfs/HQII4250.pdf}
}

@article{damato2023machine,
  author    = {Claudia d'Amato and Louis Mahon and Pierre Monnin and Giorgos Stamou},
  title     = {Machine Learning and Knowledge Graphs: Existing Gaps and Future Research Challenges},
  journal   = {Transactions on Graph Data and Knowledge},
  volume    = {1},
  number    = {1},
  pages     = {8:1--8:35},
  year      = {2023},
  publisher = {Schloss Dagstuhl -- Leibniz-Zentrum f{\"u}r Informatik},
  doi       = {10.4230/TGDK.1.1.8},
  url       = {https://doi.org/10.4230/TGDK.1.1.8}
}

@misc{jui2026lpvi,
  author       = {Jui, Shang-Ling},
  title        = {From Probabilistic {AI} to Structural Intelligence: Learn, Preserve, Verify, and Interpret: Toward a New Computational Paradigm for Scientific and Engineering Intelligence},
  year         = {2026},
  howpublished = {Chaspark, online position paper series},
  url          = {https://www.chaspark.com/#/hotspots/1336531282661879808?lang=en},
  note         = {Published in Chinese and English installments; series ongoing. URL points to the first installment}
}

@article{yu2026sharp,
  author  = {Cong Yu and Tuo Shi and Matthias Weidlich and Bo Zhao},
  title   = {{SHARP}: Shared State Reduction for Efficient Matching of Sequential Patterns},
  journal = {Proceedings of the VLDB Endowment},
  volume  = {19},
  number  = {5},
  pages   = {987--1000},
  year    = {2026},
  doi     = {10.14778/3796195.3796210}
}

@inproceedings{wagenlander2024tenplex,
  author    = {Marcel Wagenl{\"a}nder and Guo Li and Bo Zhao and Luo Mai and Peter Pietzuch},
  title     = {{Tenplex}: Dynamic Parallelism for Deep Learning using Parallelizable Tensor Collections},
  booktitle = {Proceedings of the 29th ACM Symposium on Operating Systems Principles (SOSP)},
  pages     = {195--210},
  year      = {2024},
  doi       = {10.1145/3694715.3695975}
}

@inproceedings{tan2026earl,
  author    = {Zheyue Tan and Tuo Shi and Huining Yuan and Zelai Xu and Chao Yu and Boxun Li and Yu Wang and Bo Zhao},
  title     = {{EARL}: Efficient Agentic {RL} Post-Training for {LLMs} under Dynamic Context Lengths},
  booktitle = {Proceedings of the 6th European Workshop on Machine Learning and Systems (EuroMLSys)},
  pages     = {41--48},
  year      = {2026},
  doi       = {10.1145/3805621.3807632}
}

@misc{tsoumakas2,
  author        = {Despina Christou and Grigorios Tsoumakas},
  title         = {Sub-Billion, Super-Frontier: Small Language Models Rival Zero-Shot Frontier {LLMs} on General and Literary Relation Extraction},
  year          = {2026},
  eprint        = {2606.22606},
  archivePrefix = {arXiv},
  primaryClass  = {cs.CL},
  howpublished  = {arXiv:2606.22606},
  url           = {https://arxiv.org/abs/2606.22606}
}

@inproceedings{tsoumakas3,
  author    = {Dimitris Dimitriadis and Vasiliki Patsiou and Eleonora Stoikopoulou and Achilleas Toumpas and Alexandra Bekiaridou and Athanasios Samaras and George Giannakoulas and Grigorios Tsoumakas},
  title     = {Overview of {ElCardioCC} Task on Clinical Coding in Cardiology at {BioASQ} 2026},
  booktitle = {Working Notes of {CLEF} 2026 -- Conference and Labs of the Evaluation Forum},
  series    = {CEUR Workshop Proceedings},
  volume    = {4283},
  pages     = {75--86},
  year      = {2026},
  url       = {https://ceur-ws.org/Vol-4283/}
}

@inproceedings{tsoumakas4,
  author        = {Maria Nefeli Paraskevopoulou and Tatiana Passali and Grigorios Tsoumakas},
  title         = {Less is More: Quality-Aware Training Data Selection for Scientific Summarization},
  booktitle     = {Findings of the Association for Computational Linguistics: {EACL} 2027},
  year          = {2027},
  note          = {To appear. Preprint available at arXiv:2606.24828},
  eprint        = {2606.24828},
  archivePrefix = {arXiv},
  primaryClass  = {cs.CL}
}

@unpublished{tsoumakas1,
  author = {Kots, Ege and Passali, Tatiana and Martone, Francesco and Izzo, Carmelo and Tsoumakas, Grigorios},
  title  = {A Tool-Augmented Agent for Wind Tunnel Data Analysis},
  year   = {2026},
  note   = {Manuscript submitted for publication. Available on OpenReview: \url{https://openreview.net/forum?id=I80Fm4FmCx}}
}

@book{fauconnier2002way,
  author    = {Gilles Fauconnier and Mark Turner},
  title     = {The Way We Think: Conceptual Blending and the Mind's Hidden Complexities},
  publisher = {Basic Books},
  address   = {New York},
  year      = {2002}
}

@article{meghini2018fol,
  author  = {Carlo Meghini and Martin Doerr},
  title   = {A First-Order Logic Expression of the {CIDOC} Conceptual Reference Model},
  journal = {International Journal of Metadata, Semantics and Ontologies},
  volume  = {13},
  number  = {2},
  pages   = {131--149},
  year    = {2018},
  doi     = {10.1504/IJMSO.2018.098393}
}

@incollection{reiter1978cwa,
  author    = {Raymond Reiter},
  title     = {On Closed World Data Bases},
  booktitle = {Logic and Data Bases},
  editor    = {Herv{\'e} Gallaire and Jack Minker},
  publisher = {Plenum Press},
  pages     = {55--76},
  year      = {1978}
}

@inproceedings{arnaout2021negative,
  author    = {Hiba Arnaout and Simon Razniewski and Gerhard Weikum and Jeff Z. Pan},
  title     = {Negative Knowledge for Open-world {Wikidata}},
  booktitle = {Companion Proceedings of the Web Conference 2021},
  pages     = {544--551},
  year      = {2021}
}

@article{Doerr2025IdentifiableIndividuals,
  author    = {Martin Doerr},
  title     = {Identifiable Individuals and Reality: Describing the Past by Formal Propositions},
  journal   = {ACM Journal on Computing and Cultural Heritage},
  volume    = {18},
  number    = {2},
  articleno = {27},
  pages     = {1--32},
  year      = {2025},
  doi       = {10.1145/3715011},
  url       = {https://doi.org/10.1145/3715011}
}

@article{elazar2021consistency,
  author  = {Yanai Elazar and Nora Kassner and Shauli Ravfogel and Abhilasha Ravichander and Eduard Hovy and Hinrich Sch{\"u}tze and Yoav Goldberg},
  title   = {Measuring and Improving Consistency in Pretrained Language Models},
  journal = {Transactions of the Association for Computational Linguistics},
  volume  = {9},
  pages   = {1012--1031},
  year    = {2021}
}

@inproceedings{Skaperdas2026OGNSD,
  author    = {Skaperdas, E. and Cami, K. and Bassiliades, N.},
  title     = {{OG-NSD}: Neuro-Symbolic Ontology Drafting from Natural-Language Requirements},
  booktitle = {Proceedings of the   Workshop on LLM-driven Knowledge Graph and Ontology Engineering (LLMS4KGOE 2026)},
  series    = {CEUR Workshop Proceedings},
  volume    = {4246},
  year      = {2026}
}

@inproceedings{Skaperdas2026EurostatKG,
  author    = {Skaperdas, E. and Bassiliades, N.},
  title     = {Ontology-Aware Relation Prediction over {EurostatKG}: A Unified Neural Framework},
  booktitle = {Proceedings of the International Joint Conference on Knowledge Graphs (IJCKG 2025)},
  series    = {Lecture Notes in Artificial Intelligence},
  volume    = {16297},
  publisher = {Springer Nature},
  year      = {2026}
}

@inproceedings{SimitsisSV23,
  author    = {Alkis Simitsis and Spiros Skiadopoulos and Panos Vassiliadis},
  editor    = {Enrico Gallinucci and Lukasz Golab},
  title     = {The History, Present, and Future of {ETL} Technology (invited)},
  booktitle = {Proceedings of the 25th International Workshop on Design, Optimization, Languages and Analytical Processing of Big Data ({DOLAP}), co-located with {EDBT/ICDT} 2023},
  address   = {Ioannina, Greece},
  series    = {CEUR Workshop Proceedings},
  volume    = {3369},
  pages     = {3--12},
  publisher = {CEUR-WS.org},
  year      = {2023},
  url       = {https://ceur-ws.org/Vol-3369/invited1.pdf}
}

@inproceedings{marketakis2026exploiting,
  author    = {Yannis Marketakis and Yannis Tzitzikas},
  title     = {Exploiting Mapping Corpora and Expert Knowledge for {LLM}},
  booktitle = {Joint Proceedings of the {GLOW@ISWC} 2026 Workshops: 1st International Workshop on Graph-enhanced {LLMs} for Trustworthy Web Data Management ({GLOW} 2026)},
  address   = {Bari, Italy},
  month     = oct,
  publisher = {CEUR-WS.org},
  year      = {2026}
}

@article{marketakis2017x3ml,
  author    = {Yannis Marketakis and Nikos Minadakis and Haridimos Kondylakis and Konstantina Konsolaki and Georgios Samaritakis and Maria Theodoridou and Giorgos Flouris and Martin Doerr},
  title     = {{X3ML} Mapping Framework for Information Integration in Cultural Heritage and Beyond},
  journal   = {International Journal on Digital Libraries},
  volume    = {18},
  number    = {4},
  pages     = {301--319},
  year      = {2017},
  publisher = {Springer}
}

@article{triantafillou2004cache,
  author  = {Peter Triantafillou and Nikos Ntarmos and John Yannakopoulos},
  title   = {A Cache Engine for E-Content Integration},
  journal = {IEEE Internet Computing},
  volume  = {8},
  number  = {2},
  pages   = {45--53},
  year    = {2004},
  doi     = {10.1109/MIC.2004.1273485}
}

@article{ntarmos2014rankjoin,
  author  = {Nikos Ntarmos and Ioannis Patlakas and Peter Triantafillou},
  title   = {Rank Join Queries in {NoSQL} Databases},
  journal = {Proceedings of the VLDB Endowment},
  volume  = {7},
  number  = {7},
  pages   = {493--504},
  year    = {2014}
}

@article{gruber1993,
  author  = {Thomas R. Gruber},
  title   = {A Translation Approach to Portable Ontology Specifications},
  journal = {Knowledge Acquisition},
  volume  = {5},
  number  = {2},
  pages   = {199--220},
  year    = {1993}
}

@article{hogan2021,
  author  = {Aidan Hogan and Eva Blomqvist and Michael Cochez and Claudia d'Amato and Gerard de Melo and Claudio Gutierrez and Sabrina Kirrane and Jos{\'e} Emilio Labra Gayo and Roberto Navigli and Sebastian Neumaier and Axel-Cyrille Ngonga Ngomo and Axel Polleres and Sabbir M. Rashid and Anisa Rula and Lukas Schmelzeisen and Juan Sequeda and Steffen Staab and Antoine Zimmermann},
  title   = {Knowledge Graphs},
  journal = {ACM Computing Surveys},
  volume  = {54},
  number  = {4},
  year    = {2021}
}

@article{doerr2003,
  author  = {Martin Doerr},
  title   = {The {CIDOC} Conceptual Reference Module: An Ontological Approach to Semantic Interoperability of Metadata},
  journal = {AI Magazine},
  volume  = {24},
  number  = {3},
  pages   = {75--92},
  year    = {2003}
}

@incollection{poggi2008,
  author    = {Antonella Poggi and Domenico Lembo and Diego Calvanese and Giuseppe De Giacomo and Maurizio Lenzerini and Riccardo Rosati},
  title     = {Linking Data to Ontologies},
  booktitle = {Journal on Data Semantics X},
  series    = {LNCS},
  volume    = {4900},
  pages     = {133--173},
  publisher = {Springer},
  year      = {2008}
}

@inproceedings{xiao2018,
  author    = {Guohui Xiao and Diego Calvanese and Roman Kontchakov and Domenico Lembo and Antonella Poggi and Riccardo Rosati and Michael Zakharyaschev},
  title     = {Ontology-Based Data Access: A Survey},
  booktitle = {Proceedings of the 27th International Joint Conference on Artificial Intelligence ({IJCAI})},
  pages     = {5511--5519},
  year      = {2018}
}

@article{calvanese2017ontop,
  author  = {Diego Calvanese and Benjamin Cogrel and Sarah Komla-Ebri and Roman Kontchakov and Davide Lanti and Martin Rezk and Mariano Rodriguez-Muro and Guohui Xiao},
  title   = {{Ontop}: Answering {SPARQL} Queries over Relational Databases},
  journal = {Semantic Web},
  volume  = {8},
  number  = {3},
  pages   = {471--487},
  year    = {2017}
}

@article{mountantonakis2019,
  author  = {Michalis Mountantonakis and Yannis Tzitzikas},
  title   = {Large-scale Semantic Integration of Linked Data: A Survey},
  journal = {ACM Computing Surveys},
  volume  = {52},
  number  = {5},
  year    = {2019}
}

@article{tzitzikas2017faceted,
  author  = {Yannis Tzitzikas and Nikos Manolis and Panagiotis Papadakos},
  title   = {Faceted Exploration of {RDF/S} Datasets: A Survey},
  journal = {Journal of Intelligent Information Systems},
  volume  = {48},
  number  = {2},
  pages   = {329--364},
  year    = {2017}
}

@article{mountantonakis2024text2cidoc,
  author    = {Michalis Mountantonakis and Yannis Tzitzikas},
  title     = {Generating {SPARQL} Queries over {CIDOC-CRM} Using a Two-Stage Ontology Path Patterns Method in {LLM} Prompts},
  journal   = {ACM Journal on Computing and Cultural Heritage},
  volume    = {18},
  number    = {1},
  pages     = {1--20},
  year      = {2025},
  publisher = {ACM},
  doi       = {10.1145/3708326}
}

@article{flouris2008,
  author  = {Giorgos Flouris and Dimitris Manakanatas and Haridimos Kondylakis and Dimitris Plexousakis and Grigoris Antoniou},
  title   = {Ontology Change: Classification and Survey},
  journal = {The Knowledge Engineering Review},
  volume  = {23},
  number  = {2},
  pages   = {117--152},
  year    = {2008}
}

@article{agm1985,
  author  = {Carlos E. Alchourr{\'o}n and Peter G{\"a}rdenfors and David Makinson},
  title   = {On the Logic of Theory Change: Partial Meet Contraction and Revision Functions},
  journal = {The Journal of Symbolic Logic},
  volume  = {50},
  number  = {2},
  pages   = {510--530},
  year    = {1985}
}

@article{pan2023tgdk,
  author  = {Jeff Z. Pan and Simon Razniewski and Jan-Christoph Kalo and Sneha Singhania and Jiaoyan Chen and Stefan Dietze and Hajira Jabeen and Janna Omeliyanenko and Wen Zhang and Matteo Lissandrini and Russa Biswas and Gerard de Melo and Angela Bonifati and Edlira Vakaj and Mauro Dragoni and Damien Graux},
  title   = {Large Language Models and Knowledge Graphs: Opportunities and Challenges},
  journal = {Transactions on Graph Data and Knowledge},
  volume  = {1},
  number  = {1},
  year    = {2023}
}

@article{pan2024roadmap,
  author  = {Shirui Pan and Linhao Luo and Yufei Wang and Chen Chen and Jiapu Wang and Xindong Wu},
  title   = {Unifying Large Language Models and Knowledge Graphs: A Roadmap},
  journal = {IEEE Transactions on Knowledge and Data Engineering},
  volume  = {36},
  number  = {7},
  pages   = {3580--3599},
  year    = {2024}
}

@article{damato2020,
  author  = {Claudia d'Amato},
  title   = {Machine Learning for the Semantic Web: Lessons Learnt and Next Research Directions},
  journal = {Semantic Web},
  volume  = {11},
  number  = {1},
  pages   = {195--203},
  year    = {2020}
}

@inproceedings{bordes2013,
  author    = {Antoine Bordes and Nicolas Usunier and Alberto Garc{\'i}a-Dur{\'a}n and Jason Weston and Oksana Yakhnenko},
  title     = {Translating Embeddings for Modeling Multi-relational Data},
  booktitle = {Advances in Neural Information Processing Systems 26 (NIPS 2013)},
  year      = {2013}
}

@article{tsatsaronis2015,
  author  = {George Tsatsaronis and Georgios Balikas and Prodromos Malakasiotis and Ioannis Partalas and Matthias Zschunke and Michael R. Alvers and Dirk Weissenborn and Anastasia Krithara and Sergios Petridis and Dimitris Polychronopoulos and Yannis Almirantis and John Pavlopoulos and Nicolas Baskiotis and Patrick Gallinari and Thierry Arti{\`e}res and Axel-Cyrille Ngonga Ngomo and Norman Heino and Eric Gaussier and Liliana Barrio-Alvers and Michael Schroeder and Ion Androutsopoulos and Georgios Paliouras},
  title   = {An Overview of the {BIOASQ} Large-Scale Biomedical Semantic Indexing and Question Answering Competition},
  journal = {BMC Bioinformatics},
  volume  = {16},
  pages   = {138},
  year    = {2015}
}

@inproceedings{lewis2020rag,
  author    = {Patrick Lewis and Ethan Perez and Aleksandra Piktus and Fabio Petroni and Vladimir Karpukhin and Naman Goyal and Heinrich K{\"u}ttler and Mike Lewis and Wen-tau Yih and Tim Rockt{\"a}schel and Sebastian Riedel and Douwe Kiela},
  title     = {Retrieval-Augmented Generation for Knowledge-Intensive {NLP} Tasks},
  booktitle = {Advances in Neural Information Processing Systems 33 (NeurIPS 2020)},
  year      = {2020}
}

@misc{edge2024graphrag,
  author       = {Darren Edge and Ha Trinh and Newman Cheng and Joshua Bradley and Alex Chao and Apurva Mody and Steven Truitt and Jonathan Larson},
  title        = {From Local to Global: A Graph {RAG} Approach to Query-Focused Summarization},
  howpublished = {arXiv:2404.16130},
  year         = {2024}
}

@inproceedings{yao2023react,
  author    = {Shunyu Yao and Jeffrey Zhao and Dian Yu and Nan Du and Izhak Shafran and Karthik Narasimhan and Yuan Cao},
  title     = {{ReAct}: Synergizing Reasoning and Acting in Language Models},
  booktitle = {International Conference on Learning Representations (ICLR)},
  year      = {2023}
}

@article{sumers2024coala,
  author  = {Theodore R. Sumers and Shunyu Yao and Karthik Narasimhan and Thomas L. Griffiths},
  title   = {Cognitive Architectures for Language Agents},
  journal = {Transactions on Machine Learning Research},
  year    = {2024}
}

@misc{packer2023memgpt,
  author       = {Charles Packer and Sarah Wooders and Kevin Lin and Vivian Fang and Shishir G. Patil and Ion Stoica and Joseph E. Gonzalez},
  title        = {{MemGPT}: Towards {LLMs} as Operating Systems},
  howpublished = {arXiv:2310.08560},
  year         = {2023}
}

@inproceedings{park2023generative,
  author    = {Joon Sung Park and Joseph C. O'Brien and Carrie J. Cai and Meredith Ringel Morris and Percy Liang and Michael S. Bernstein},
  title     = {Generative Agents: Interactive Simulacra of Human Behavior},
  booktitle = {Proceedings of the 36th Annual ACM Symposium on User Interface Software and Technology (UIST)},
  year      = {2023}
}

@article{liu2024lost,
  author  = {Nelson F. Liu and Kevin Lin and John Hewitt and Ashwin Paranjape and Michele Bevilacqua and Fabio Petroni and Percy Liang},
  title   = {Lost in the Middle: How Language Models Use Long Contexts},
  journal = {Transactions of the Association for Computational Linguistics},
  volume  = {12},
  pages   = {157--173},
  year    = {2024}
}

@inproceedings{kwon2023vllm,
  author    = {Woosuk Kwon and Zhuohan Li and Siyuan Zhuang and Ying Sheng and Lianmin Zheng and Cody Hao Yu and Joseph E. Gonzalez and Hao Zhang and Ion Stoica},
  title     = {Efficient Memory Management for Large Language Model Serving with {PagedAttention}},
  booktitle = {Proceedings of the 29th Symposium on Operating Systems Principles (SOSP)},
  year      = {2023}
}

@article{ji2023hallucination,
  author  = {Ziwei Ji and Nayeon Lee and Rita Frieske and Tiezheng Yu and Dan Su and Yan Xu and Etsuko Ishii and Ye Jin Bang and Andrea Madotto and Pascale Fung},
  title   = {Survey of Hallucination in Natural Language Generation},
  journal = {ACM Computing Surveys},
  volume  = {55},
  number  = {12},
  year    = {2023}
}

@misc{sifakis2018autonomous,
  author       = {Joseph Sifakis},
  title        = {Autonomous Systems -- An Architectural Characterization},
  howpublished = {arXiv:1811.10277},
  year         = {2018}
}

@article{harel2020autonomics,
  author  = {David Harel and Assaf Marron and Joseph Sifakis},
  title   = {Autonomics: In Search of a Foundation for Next-Generation Autonomous Systems},
  journal = {Proceedings of the National Academy of Sciences},
  volume  = {117},
  number  = {30},
  pages   = {17491--17498},
  year    = {2020}
}

@article{clarke2009,
  author  = {Edmund M. Clarke and E. Allen Emerson and Joseph Sifakis},
  title   = {Model Checking: Algorithmic Verification and Debugging},
  journal = {Communications of the ACM},
  volume  = {52},
  number  = {11},
  pages   = {74--84},
  year    = {2009}
}

@article{saltzer1975,
  author  = {Jerome H. Saltzer and Michael D. Schroeder},
  title   = {The Protection of Information in Computer Systems},
  journal = {Proceedings of the IEEE},
  volume  = {63},
  number  = {9},
  pages   = {1278--1308},
  year    = {1975}
}

@book{gray1992,
  author    = {Jim Gray and Andreas Reuter},
  title     = {Transaction Processing: Concepts and Techniques},
  publisher = {Morgan Kaufmann},
  year      = {1992}
}

@misc{prov2013,
  author       = {Timothy Lebo and Satya Sahoo and Deborah McGuinness},
  title        = {{PROV-O}: The {PROV} Ontology},
  howpublished = {W3C Recommendation},
  year         = {2013}
}

@misc{shacl2017,
  author       = {Holger Knublauch and Dimitris Kontokostas},
  title        = {Shapes Constraint Language ({SHACL})},
  howpublished = {W3C Recommendation},
  year         = {2017}
}

@article{kirkpatrick2017,
  author  = {James Kirkpatrick and Razvan Pascanu and Neil Rabinowitz and Joel Veness and Guillaume Desjardins and Andrei A. Rusu and Kieran Milan and John Quan and Tiago Ramalho and Agnieszka Grabska-Barwinska and Demis Hassabis and Claudia Clopath and Dharshan Kumaran and Raia Hadsell},
  title   = {Overcoming Catastrophic Forgetting in Neural Networks},
  journal = {Proceedings of the National Academy of Sciences},
  volume  = {114},
  number  = {13},
  pages   = {3521--3526},
  year    = {2017}
}

@article{vassiliadis2019intentional,
  author  = {Panos Vassiliadis and Patrick Marcel and Stefano Rizzi},
  title   = {Beyond Roll-Up's and Drill-Down's: An Intentional Analytics Model to Reinvent {OLAP}},
  journal = {Information Systems},
  volume  = {85},
  pages   = {68--91},
  year    = {2019}
}

@article{tsagkatakis2019survey,
  author  = {Grigorios Tsagkatakis and Anastasia Aidini and Konstantina Fotiadou and Michalis Giannopoulos and Anastasia Pentari and Panagiotis Tsakalides},
  title   = {Survey of Deep-Learning Approaches for Remote Sensing Observation Enhancement},
  journal = {Sensors},
  volume  = {19},
  number  = {18},
  pages   = {3929},
  year    = {2019}
}

@article{ioannidis1996qo,
  author  = {Yannis E. Ioannidis},
  title   = {Query Optimization},
  journal = {ACM Computing Surveys},
  volume  = {28},
  number  = {1},
  pages   = {121--123},
  year    = {1996}
}

@inproceedings{ioannidis2003histograms,
  author    = {Yannis E. Ioannidis},
  title     = {The History of Histograms (abridged)},
  booktitle = {Proceedings of the 29th International Conference on Very Large Data Bases ({VLDB})},
  pages     = {19--30},
  year      = {2003}
}

@book{pan2017linked,
  editor    = {Jeff Z. Pan and Guido Vetere and Jos{\'e} Manuel G{\'o}mez-P{\'e}rez and Honghan Wu},
  title     = {Exploiting Linked Data and Knowledge Graphs in Large Organisations},
  publisher = {Springer},
  year      = {2017}
}

@article{khan2023kgquerying,
  author  = {Arijit Khan},
  title   = {Knowledge Graphs Querying},
  journal = {ACM SIGMOD Record},
  volume  = {52},
  number  = {2},
  pages   = {18--29},
  year    = {2023}
}

@inproceedings{collarana2017minte,
  author    = {Diego Collarana and Mikhail Galkin and Ignacio Traverso-Rib{\'o}n and Maria-Esther Vidal and Christoph Lange and S{\"o}ren Auer},
  title     = {{MINTE}: Semantically Integrating {RDF} Graphs},
  booktitle = {Proceedings of the 7th International Conference on Web Intelligence, Mining and Semantics ({WIMS})},
  year      = {2017}
}

@inproceedings{vassiliadis2002etl,
  author    = {Panos Vassiliadis and Alkis Simitsis and Spiros Skiadopoulos},
  title     = {Conceptual Modeling for {ETL} Processes},
  booktitle = {Proceedings of the 5th ACM International Workshop on Data Warehousing and {OLAP} ({DOLAP})},
  pages     = {14--21},
  year      = {2002}
}

@book{chalkiadakis2011cgt,
  author    = {Georgios Chalkiadakis and Edith Elkind and Michael Wooldridge},
  title     = {Computational Aspects of Cooperative Game Theory},
  series    = {Synthesis Lectures on Artificial Intelligence and Machine Learning},
  publisher = {Morgan \& Claypool},
  year      = {2011}
}

@article{tsoumakas2007multilabel,
  author  = {Grigorios Tsoumakas and Ioannis Katakis},
  title   = {Multi-Label Classification: An Overview},
  journal = {International Journal of Data Warehousing and Mining},
  volume  = {3},
  number  = {3},
  pages   = {1--13},
  year    = {2007}
}

@misc{t_sifakis, author = {Joseph Sifakis}, title = {Trustworthy Autonomous Agents}, howpublished = {Keynote, Olympia AI Data Infrastructure Summit 2026, Heraklion, Greece}, year = {2026}}

@misc{t_calvanese, author = {Diego Calvanese}, title = {Knowledge Graphs from an Industrial Perspective}, howpublished = {Keynote, Olympia AI Data Infrastructure Summit 2026, Heraklion, Greece}, year = {2026}}

@misc{t_ioannidis, author = {Yannis Ioannidis}, title = {The Future of Data Infrastructure in the Agentic {AI} Era}, howpublished = {Keynote, Olympia AI Data Infrastructure Summit 2026, Heraklion, Greece}, year = {2026}}

@misc{t_pan, author = {Jeff Z. Pan}, title = {From Parameter Scaling to Agent Scaling}, howpublished = {Invited talk, Olympia AI Data Infrastructure Summit 2026, Heraklion, Greece}, year = {2026}}

@misc{t_zhao, author = {Bo Zhao}, title = {State Management for Agentic {AI} Systems}, howpublished = {Invited talk, Olympia AI Data Infrastructure Summit 2026, Heraklion, Greece}, year = {2026}}

@misc{t_doerr, author = {Martin Doerr}, title = {Challenges of Knowledge about the Past}, howpublished = {Invited talk, Olympia AI Data Infrastructure Summit 2026, Heraklion, Greece}, year = {2026}}

@misc{t_damato, author = {Claudia d'Amato}, title = {Neural-Symbolic Link Prediction Methods, Benchmarks and Explanations for Knowledge Graph Refinement}, howpublished = {Invited talk, Olympia AI Data Infrastructure Summit 2026, Heraklion, Greece}, year = {2026}}

@misc{t_bassiliades, author = {Nick Bassiliades}, title = {Bridging Language Models and Semantic Knowledge: Neuro-Symbolic Approaches to Ontology Engineering, Alignment, and Knowledge Graph Completion}, howpublished = {Invited talk, Olympia AI Data Infrastructure Summit 2026, Heraklion, Greece}, year = {2026}}

@misc{t_collarana, author = {Diego Collarana}, title = {Beyond Words: Grounding {AI} Agents in Enterprise Knowledge}, howpublished = {Invited talk, Olympia AI Data Infrastructure Summit 2026, Heraklion, Greece}, year = {2026}}

@misc{t_paliouras, author = {Georgios Paliouras}, title = {The {BioASQ} Challenge: Resources and Evaluation for Semantic Extraction from the Biomedical Literature}, howpublished = {Invited talk, Olympia AI Data Infrastructure Summit 2026, Heraklion, Greece}, year = {2026}}

@misc{t_tsoumakas, author = {Grigoris Tsoumakas}, title = {From Semantic Extraction to Autonomous Agents: Architecting the Next Generation of Enterprise {AI}}, howpublished = {Invited talk, Olympia AI Data Infrastructure Summit 2026, Heraklion, Greece}, year = {2026}}

@misc{t_khan, author = {Arijit Khan}, title = {Towards Graph-native {LLMs} and Agentic Systems: Rethinking the {AI} Harness}, howpublished = {Invited talk, Olympia AI Data Infrastructure Summit 2026, Heraklion, Greece}, year = {2026}}

@misc{t_tsagkatakis, author = {Grigoris Tsagkatakis}, title = {Agentic Frameworks for Space Data Management: Semantic Access, Agent Memory, and Scaling to Operational Services}, howpublished = {Invited talk, Olympia AI Data Infrastructure Summit 2026, Heraklion, Greece}, year = {2026}}

@misc{t_vassiliadis, author = {Panos Vassiliadis}, title = {From Analytic Intentions to Data Stories via Highlights}, howpublished = {Invited talk, Olympia AI Data Infrastructure Summit 2026, Heraklion, Greece}, year = {2026}}

@misc{t_magoutis, author = {Kostas Magoutis}, title = {Autoscaling of {AI} Services}, howpublished = {Invited talk, Olympia AI Data Infrastructure Summit 2026, Heraklion, Greece}, year = {2026}}

@misc{t_pitas, author = {Ioannis Pitas}, title = {Decentralized Machine Learning}, howpublished = {Invited talk, Olympia AI Data Infrastructure Summit 2026, Heraklion, Greece}, year = {2026}}

@misc{t_chalkiadakis, author = {Georgios Chalkiadakis}, title = {[Safe] Uncertainty Driven Adaptive Exploration: Towards Enabling Agent Self-Evolution through Experience}, howpublished = {Invited talk, Olympia AI Data Infrastructure Summit 2026, Heraklion, Greece}, year = {2026}}

@misc{t_jui, author = {Shang-Ling Jui}, title = {From Data Infrastructure to Structural Intelligence: Ontology, Memory and Verifiable Reasoning for the Next Generation of {AI} Agents}, howpublished = {Invited talk, Olympia AI Data Infrastructure Summit 2026, Heraklion, Greece}, year = {2026}}

@misc{t_panel, author = {Yannis Ioannidis and Diego Calvanese and Dimitris Plexousakis and Arijit Khan and Claudia d'Amato}, title = {Ontology and Agent Memory: Turning Research into Product Capabilities}, howpublished = {Panel moderated by Y. Tzitzikas, Olympia AI Data Infrastructure Summit 2026, Heraklion, Greece}, year = {2026}}

\end{document}